\documentclass{article}
\usepackage{ijcai26}

\usepackage{times}
\usepackage{soul}
\usepackage{url}
\usepackage[hidelinks]{hyperref}
\usepackage[utf8]{inputenc}
\usepackage[small]{caption}
\usepackage{graphicx}
\usepackage{amsmath}
\usepackage{amsthm}
\usepackage{booktabs}
\usepackage{algorithm}
\usepackage{algorithmic}
\usepackage[switch]{lineno}
\usepackage{bm}
\usepackage{color}
\usepackage{xcolor}
\usepackage{array}
\usepackage{colortbl}
\usepackage{multirow}
\usepackage{mathrsfs}
\usepackage{amssymb}

\definecolor{codeblue}{RGB}{0, 82, 147}     
\definecolor{codegreen}{RGB}{0, 128, 0}    
\definecolor{codegray}{RGB}{100, 100, 100}  
\definecolor{codeorange}{RGB}{230, 145, 56} 
\definecolor{darkerblue}{rgb}{0,0.08,0.45}
\definecolor{royalblue}{RGB}{65,105,225}
\definecolor{lightblue}{RGB}{221,235,247}
\definecolor{figblue}{RGB}{47, 122, 232}  
\definecolor{figred}{RGB}{213, 32, 52}
\definecolor{figgreen}{RGB}{0, 137, 72} 
\definecolor{figyellow}{RGB}{217, 161, 5}
\definecolor{gray94}{gray}{.94}
\definecolor{gray90}{gray}{.90}

\newcommand{\blue}[1]{\textcolor{darkerblue}{#1}}
\definecolor{darkgreen}{RGB}{34,139,34}
\newcommand{\green}[1]{\textcolor{darkgreen}{#1}}
\newcommand{\gray}[1]{\textcolor{codegray}{#1}}
\newcommand{\gbf}[1]{\green{\bf{#1}}}

\definecolor{darkpurple}{RGB}{110,51,137}
\newcommand{\dpbf}[1]{\textcolor{darkpurple}{\bf{#1}}}

\newcolumntype{g}{>{\columncolor{gray94}}c} 
\newcommand{\grow}[1]{\rowcolor{gray94}{#1}} 
\newcommand{\brow}[1]{\rowcolor{lightblue}{#1}} 
\usepackage{pifont}
\usepackage{latexsym}

\title{VVTRec: Radio Interferometric Reconstruction through Visual and Textual Modality Enrichment}

\author{
Kai Cheng$^1$\and
Ruoqi Wang$^2$\And
Qiong Luo\footnote{Corresponding Author.}$^{,1,2}$\\
\affiliations
$^1$The Hong Kong University of Science and Technology\\
$^2$The Hong Kong University of Science and Technology (Guangzhou)\\
\emails
kai.cheng@connect.ust.hk,
rwang280@connect.hkust-gz.edu.cn,
luo@ust.hk
}

\begin{document}

\maketitle

\begin{abstract}
    Radio astronomy is an indispensable discipline for observing distant celestial objects. Measurements of wave signals from radio telescopes, called visibility, need to be transformed into images for astronomical observations. These dirty images blend information from real sources and artifacts. Therefore, astronomers usually perform reconstruction before imaging to obtain cleaner images. Existing methods consider only a single modality of sparse visibility data, resulting in images with remaining artifacts and insufficient modeling of correlation. To enhance the extraction of visibility information and emphasize output quality in the image domain, we propose VVTRec, a multimodal radio interferometric data reconstruction method with visibility-guided visual and textual modality enrichment. In our VVTRec, sparse visibility is transformed into image-form and text-form features to obtain enhancements in terms of spatial and semantic information, improving the structural integrity and accuracy of images. Also, we leverage Vision-Language Models (VLMs) to achieve additional training-free performance improvements. VVTRec enables sparse visibility, as a foreign modality unseen by VLMs, to accurately extract pre-trained knowledge as a supplement. Our experiments demonstrate that VVTRec effectively enhances imaging results by exploiting multimodal information without introducing excessive computational overhead.
\end{abstract}

\section{Introduction}

\begin{figure}[ht]
    \centering
    \includegraphics[width=.85\columnwidth]{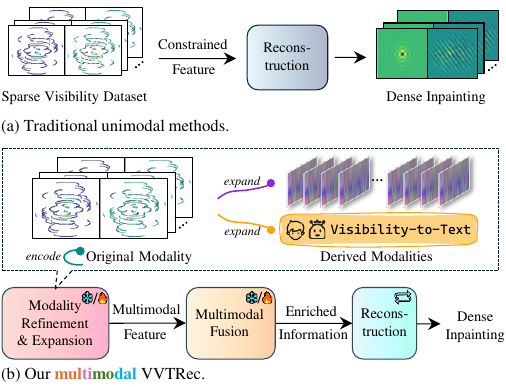}
    \caption{Comparison of traditional unimodal visibility data processing paradigm and our multimodal pipeline.}
    \label{fig:fig1}
\end{figure}

In radio astronomy, \textit{visibility} refers to the complex data measured in the \textit{uv-plane} by radio interferometers, representing the cross-correlation between pairs of antenna signals. It characterizes the distribution of the sky in the frequency domain and serves as the source for interferometric imaging. Images directly transformed from such visibility data are called \textit{dirty images}, because they differ significantly from the real sky \cite{hogbom1974aperture,ables1974maximum,bouman2016computational,connor2022deep}. This discrepancy arises due to artifacts and various other factors \cite{schmidt2022deep}. On the one hand, visibility data are inherently sparse. Due to limitations in telescope configurations, the data are distributed only along a small number of trajectories, leaving most spatial frequency regions unmeasured. On the other hand, phenomena such as cosmic microwave background, galactic radiation, atmospheric disturbances, and receiver interferences introduce additional radio noise \cite{wilson1979cosmic,singh2022detection}. Consequently, visibility data typically require reconstruction before subsequent scientific studies. In this paper, we propose a reconstruction method that enhances imaging quality by exploiting multimodal information.

The first challenge in visibility data reconstruction lies in the inherently limited information in sparse visibility. As shown in Figure~\ref{fig:fig1}(a), sparse visibility is a weak signal whose distribution on the uv-plane is non-uniform and exhibits certain discreteness. This sparsity and context insufficiency render constrained features. However, downstream reconstructions in previous unimodal methods solely rely on the information provided by visibility, which makes it challenging to predict as many blank spatial regions as possible while maintaining prediction accuracy based on a small number of known signal locations.

The second challenge is that current strategies can only exploit unimodal complex-valued visibility, leading to some defects in reconstructed images. For example, structural information about images or semantic context cannot be adequately derived from this single modality of sparse visibility alone. Furthermore, relying solely on unimodal sparse visibility underestimates the interplay between the frequency and final target image domains \cite{zhao2025public}, reducing the reconstruction quality. Moreover, previous methods often ignore explicit spatial modeling in the unimodal setting \cite{schmidt2022deep,wang2024polarrec,wang2025visrec}, leading to further limitations in their ability to extract spatial correlations. This makes a relatively negative impact on final imaging results.

To overcome these challenges, we propose VVTRec, a radio interferometric data reconstruction approach that turns sparse astronomical observations into visibility-guided visual and textual modality enrichment. Leveraging external knowledge as a supplement is beneficial, further necessitating multimodal enhancements. As shown in Figure~\ref{fig:fig1} (b), through modality expansion, our VVTRec designs transformations based on the characteristics of sparse visibility, converting visibility into image-form and text-form modalities. The image-form modality is a generated feature map of visibility data to explicitly enhance spatial information. The text-form modality provides dataset-level and sample-level descriptions of visibility data to supplement semantic information. The three modalities collectively provide multiple sources for reconstruction, enabling access to richer underlying information and pretrained knowledge.

Both derived modalities are designed to address the issue of integrating external knowledge since visibility is foreign to pretrained models. Based on the perspective of the belief network, we propose a knowledge bank that uses encoded visibility as the query to perform the integration and selection of multimodal external knowledge in a sample-specific manner. The generated image-form modality preserves key features from visibility while also introducing an image structure. Specifically, previous methods involve the image domain only in the final imaging step, whereas our VVTRec establishes connections between the two domains at the beginning of the process. Meanwhile, the text-form modality also provides rich textual descriptions, enabling visibility to extract contextual features.

By leveraging the pre-trained knowledge of Visual-Language Models (VLMs) \cite{ZhangGHCZ025,ZhongR0LW025,ZhaoD0GB0W25}, VVTRec performs joint modeling across modalities seamlessly within a unified space and further enhances the performance. It is highly compatible with current pretrained models \cite{kim2021vilt,radford2021learning,li2023blip,xue2025blip}. For sparse visibility, the pre-trained knowledge within VLMs represents two foreign modalities. Similarly, for VLMs, visibility itself is also a foreign modality. Therefore, VVTRec exploits their dependencies and enables the transformed modalities to effectively extract proper external complementary knowledge from pre-trained models. Notably, VVTRec focuses on fully utilizing existing pre-trained knowledge to achieve performance gains. Additionally, VVTRec incurs little additional computational overhead or causes speed drops sharply when incorporating more modalities. In other words, VVTRec enhances model performance while maintaining efficiency.

Overall, our contributions can be summarized as follows:

\begin{itemize}
    \item We propose VVTRec to transform sparse visibility into expanded modalities to enhance spatial and semantic information extraction, addressing challenges in radio interferometric data reconstruction.
    \item We propose a sample-specific knowledge bank to leverage and fuse external pre-trained knowledge for improved reconstruction by joint modeling of mutually foreign modalities into multimodal features.
    \item VVTRec can gain training-free performance improvements, and experimental results show that it successfully enhances reconstructions across different datasets with little additional computational overhead.
\end{itemize}

\section{Background and Related Work}
\subsection{Radio Interferometric Imaging}
Radio telescopes offer many unique advantages for observing the universe within the radio transmission windows. They can penetrate interstellar dust and reveal the internal structures \cite{steyn2024h}. Additionally, they provide insights into unique physical mechanisms \cite{xu20220,chime2022sub}. They can also form a virtual Earth-sized telescope through Very Long Baseline Interferometry (VLBI) \cite{bouman2018reconstructing}. However, limited baselines are one of the main reasons for the sparsity of visibility data \cite{thompson2017interferometry,bouman2018reconstructing}, highlighting the importance of designing algorithms for reconstruction.

While visibility data represented in the uv-plane facilitates signal processing operations, it is unsuitable for intuitive observations. By applying the Inverse Fourier Transform (IFT) \cite{jin2025multi}, frequency-domain features can be remapped back into the image domain, which can later be studied visually \cite{wang2023conditional}. The imaging process of the spatial intensity distribution is represented as follows:
\begin{equation}
    I(l, m) = \int_u \int_v \mathcal{M} \left(u,v\right) e^{j 2\pi (ul + vm)} \, du \, dv,
\end{equation}
where $\mathcal{M} \left(u,v\right)$ denotes the frequency-domain visibility data and $I(l, m)$ denotes the intensity distribution of the sky.

\begin{figure*}[t]
\centering
    \includegraphics[width=\linewidth]{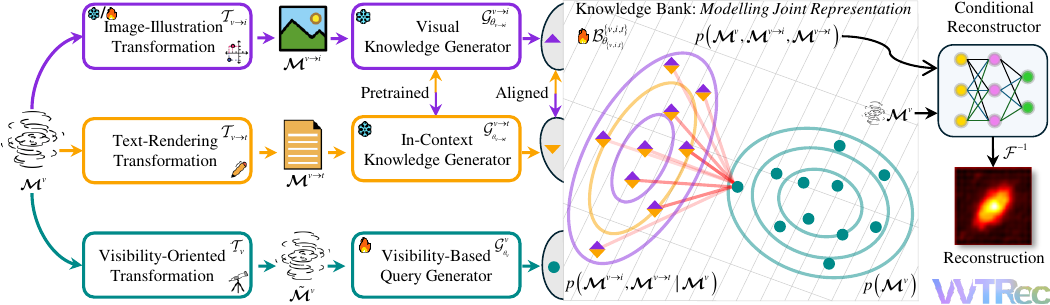}
    \caption{Overview of our proposed method VVTRec. It consists of four key components: (1) The transformations provide essential modality refinement and expansion for viewing visibility data from a multimodal perspective. (2) The generators encode rich and consistent information from three modalities into latent spaces for further alignment and extraction. (3) The knowledge bank retrieves the enriched multimodal features in a unified space by modelling the joint distribution of all modalities. (4) The conditional reconstructor is designed to combine extracted features into the reconstruction and imaging process.
    }
    \label{fig:fig2}
\end{figure*}

\subsection{Interferometric Data Reconstruction}
Radio interferometric imaging inherently constitutes an ill-posed inverse problem due to the sparse sampling in the uv-plane, while direct imaging results in severe artifacts \cite{wang2025visrec}. Traditional methods address this by iteratively removing artifacts to obtain \textit{clean images}. The CLEAN algorithm \cite{cornwell2008multiscale} iteratively removes the dirty beam in the image domain. This approach aims to overcome the ambiguity caused by incomplete data by suppressing coherent interference. However, the assumption of point-like sources in CLEAN results in unrealistic reconstructions for other types of sources \cite{connor2022deep}. Furthermore, the iterative method of CLEAN can lead to a long inference time.

Recently, some learning-based approaches have made advances by focusing on the reconstruction of visibility data. Early methods utilized dirty images to make major contributions to the conversion into clean images, regardless of the noise before imaging \cite{bouman2016computational,connor2022deep}. However, reconstructing visibility data is later demonstrated to be more effective than image reconstruction and becomes the prevailing choice. Specifically, Schmidt \textit{et al.} \cite{schmidt2022deep} introduced convolutional neural network models for reconstructing incomplete data in radio imaging. Wu \textit{et al.} \cite{wu2022neural} proposed a method to embed visibility measurements into input tokens by exploring their features in the hidden space. Wang \textit{et al.} \cite{wang2024polarrec} introduced a radial visibility loss to consider all components in the angular coordinates. Wang \textit{et al.} also \cite{wang2025visrec} explored supervised and self-supervised learning of interferometric image reconstruction. In comparison to existing work, our VVTRec is capable of effectively exploiting multimodal knowledge by integrating external knowledge to supplement sparse visibility and boost reconstruction performance.

\section{Methodology}
\subsection{Overall Architechture}
The overall framework of VVTRec is depicted in Figure~\ref{fig:fig2}. $\mathcal{M}^{v}$ denotes sparse visibility and is defined as:
\begin{equation}
\begin{split}
    \mathcal{M}^{v} = \bm{D}_{\Lambda} \mathcal{F}\left(\bm{x}\right) + \epsilon,
\end{split}
\end{equation}
where $\bm{x}$ is the corresponding image, $\mathcal{F}$ is the Fourier transformation and $\epsilon$ represents the noise. $\bm{D}_{\Lambda}$ is an under-sampling matrix where ${\Lambda}$ denotes telescopes' sampling pattern. The image-illustration transformation $\mathcal{T}_{v \rightarrow i}: \Omega_{\text{freq}} \mapsto \Omega_{\text{spatial}}$ and text-rendering transformation $\mathcal{T}_{v \rightarrow t}: \Omega_{\text{freq}} \mapsto \Omega_{\text{semantic}}$ are employed to produce the essential image-form modality $\mathcal{M}^{v \rightarrow i}$ and text-form modality $\mathcal{M}^{v \rightarrow t}$ of visibility data. A visibility-oriented transformation is also utilized to refine the encoding of visibility modality $\mathcal{M}^{v}$, outputting $\tilde{\mathcal{M}}^{v}$. On the one hand, these modalities are different in form, and each provides distinct and rich information. On the other hand, they are all rooted in the same source and collaboratively represent visibility data. Thus, we utilize generators to encode information from three modalities into latent spaces for further knowledge alignment and extraction. Specifically, $\mathcal{G}^{v \rightarrow i}_{\theta_{v \rightarrow i}}$, $\mathcal{G}^{v \rightarrow t}_{\theta_{v \rightarrow t}}$, and $\mathcal{G}^{v}_{\theta_{v}}$ denote the Visual Knowledge Generator (VKG), In-context Knowledge Generator (IKG), and Visibility Query Generator (VQG), respectively. To facilitate the smooth fusion between the vision-language domain and interferometric frequency domain, we adopt the Transformer encoder \cite{vaswani2017attention} to establish VQG. We omit the subscripts for parameters in these notations for simplicity. Subsequently, the Knowledge Bank (KB) retrieves the enriched multimodal features in a unified space by modelling the joint distribution of all modalities. The Conditional Reconstructor (CR) is designed to combine extracted features and facilitate the reconstruction and imaging. Overall, our architecture enables effective information fusion from foreign modalities to boost the radio interferometric data reconstruction. Pre-trained knowledge is therefore sufficiently utilized, and a considerable number of parameters can be frozen without introducing excessive computational overhead. Moreover, datasets used in pre-trained models \cite{kim2021vilt,radford2021learning,li2023blip,xue2025blip} do not contain content in the field of radio interferometry, further ensuring the effectiveness and generalization.

\begin{figure}[ht]
    \centering
    \includegraphics[width=.85\columnwidth]{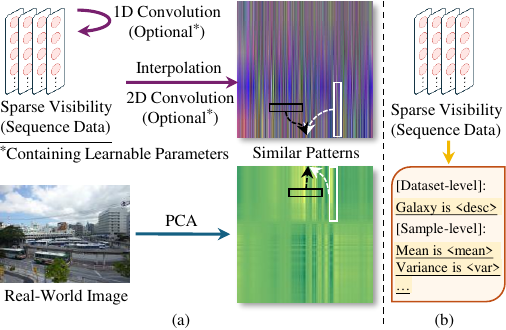}
    \caption{(a) Comparison of the generated image-form feature map and PCs of an image in the pretraining dataset \protect\cite{lin2014microsoft}. (b) An example of retrieving in-context information from sparse visibility.}
    \label{fig:fig3}
\end{figure}

\subsection{Modality Transformation}
Modelling visibility data from a multimodal view is beneficial for reconstruction from sparse visibility. Transforming visibility into more types of modalities in compliance with the input requirements of pre-trained models is vital. Visibility data are discrete signals with strong astronomical significance, which are formalized in the frequency domain. Thus, our VVTRec takes into account both the preservation of astronomical information and the transformation of modal forms. The steps of the image-illustration transformation $\mathcal{T}_{v \rightarrow i}$ can be formulated as follows:
\begin{equation}
    \mathcal{M}^{v \rightarrow i} = \operatorname{Conv}^{2D}_{\theta_{c,2}} \left( \mathcal{I} \left( \operatorname{Conv}^{1D}_{\theta_{c,1}}(\mathcal{M}^{v}) \right) \right),
\end{equation}
where $\operatorname{Conv}^{1D}$ and $\operatorname{Conv}^{2D}$ are optional 1D and 2D convolutions, respectively. $\mathcal{I}$ is the interpolation operation. The rationale behind $\mathcal{T}_{v \rightarrow i}$ is also shown in Figure~\ref{fig:fig3} (a). We sample an image of the dataset [Lin et al., 2014] used for VLM pretraining and further process it by Principal Component Analysis (PCA). The key observations are the similar patterns in our generated image-form feature map and the visualization of PCs, suggesting that VLM has the potential to extract information from sparse visibility data within our proposed framework. Moreover, to better leverage the pre-trained ability in the aligned vision-language space, we also extract the in-context information from sparse visibility as shown in Figure~\ref{fig:fig3} (b). At the dataset level, entities such as research subjects and metadata are briefly described. At the sample level, statistics are calculated as part of the prompt as well. It enables external knowledge to supplement the prior of reconstruction with semantic information.

\subsection{Enrichment from Sample-Specific Knowledge}

As shown in Figure~\ref{fig:fig2}, the visual and textual modalities $\left\{ \mathcal{M}^{v \rightarrow i},\mathcal{M}^{v \rightarrow t} \right\}$ are both derived from the visibility modality $\mathcal{M}^{v}$, implying their dependent relationships. We can formulate the goal of our KB $\mathcal{B}^{\left\{v,i,t\right\}}$ as modelling the joint representation $p\left( \mathcal{M}^{v \rightarrow i},\mathcal{M}^{v \rightarrow t},\mathcal{M}^{v} \right)$. However, directly obtaining this joint representation is challenging. We observe that a directed acyclic graph naturally formed in modality expansion can help ease the modelling if we split this joint representation as two parts from the perspective of a belief network, as follows: 
\begin{equation}
    \resizebox{.43\textwidth}{!}{$
        p\left( \mathcal{M}^{v \rightarrow i},\mathcal{M}^{v \rightarrow t},\mathcal{M}^{v} \right) = p\left( \mathcal{M}^{v \rightarrow i},\mathcal{M}^{v \rightarrow t}|\mathcal{M}^{v} \right) p\left( \mathcal{M}^{v} \right).$}
\end{equation}
Both VKG $\mathcal{G}^{v \rightarrow i}$ and IKG $\mathcal{G}^{v \rightarrow t}$ can exploit modalities $\mathcal{M}^{v \rightarrow i}$ and $\mathcal{M}^{v \rightarrow t}$ well, as they are already aligned in the feature space during the pretraining of VLM. This is another design advantage of VVTRec in addition to the training-free gain, since it can skip training on the conditional joint representation of $\mathcal{M}^{v \rightarrow i}$ and $\mathcal{M}^{v \rightarrow t}$. Thus, we define $\bm{\xi}_{\mathcal{M}^{v \rightarrow i},\mathcal{M}^{v \rightarrow t}|\mathcal{M}^{v}}$ to represent $p\left( \mathcal{M}^{v \rightarrow i},\mathcal{M}^{v \rightarrow t}|\mathcal{M}^{v} \right)$ as follows:
\begin{equation}
    \resizebox{.43\textwidth}{!}{$
        \bm{\xi}_{\mathcal{M}^{v \rightarrow i},\mathcal{M}^{v \rightarrow t}|\mathcal{M}^{v}} = \mathcal{A}\left( \mathcal{G}^{v \rightarrow i}\left(\mathcal{M}^{v \rightarrow i}\right), \mathcal{G}^{v \rightarrow t}\left(\mathcal{M}^{v \rightarrow t}\right) \right),$}
\end{equation}
where $\mathcal{A}\left(\bm{\alpha},\bm{\beta}\right)=\bm{\alpha}||\bm{\beta}$ is the concatenation operation. Moreover, we can define $\bm{\zeta}_{\mathcal{M}^{v}}$ from the output of VQG $\mathcal{G}^{v}$ to serve as the latent representation of $p\left( \mathcal{M}^{v} \right)$, which can be formulated as:
\begin{equation}
    \begin{aligned}
        \bm{\zeta}_{\mathcal{M}^{v}} = \mathcal{G}^{v}\left(\mathcal{M}^{v}\right).
    \end{aligned}
\end{equation}

\begin{figure}[b]
    \centering
    \includegraphics[width=\columnwidth]{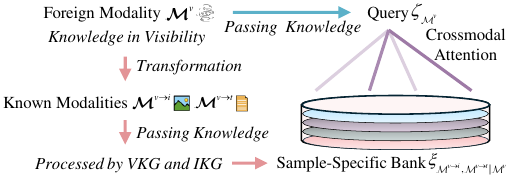}
    \caption{Illustration of the workflow of the sample-specific knowledge bank, along with the knowledge pass and integration.}
    \label{fig:fig4}
\end{figure}

Based on $\bm{\xi}_{\mathcal{M}^{v \rightarrow i},\mathcal{M}^{v \rightarrow t}|\mathcal{M}^{v}}$ and $\bm{\zeta}_{\mathcal{M}^{v}}$, we can obtain the latent representation of $p\left( \mathcal{M}^{v \rightarrow i},\mathcal{M}^{v \rightarrow t},\mathcal{M}^{v} \right)$ as follows:
\begin{equation}
    \begin{aligned}
        \bm{\eta}_{\left\{v,i,t\right\}} = \mathcal{B}^{\left\{v,i,t\right\}}\left( \bm{\xi}_{\mathcal{M}^{v \rightarrow i},\mathcal{M}^{v \rightarrow t}|\mathcal{M}^{v}},\bm{\zeta}_{\mathcal{M}^{v}} \right).
    \end{aligned}
\end{equation}
To fully capture the underlying characteristics and correlations of modalities, we design KB to obtain effective multimodal features with enriched information to facilitate reconstruction, as shown in Figure~\ref{fig:fig4}. For VKG and IKG, $\mathcal{M}^v$ is a foreign modality and inapplicable to direct process. After VVTRec employs the transformations, VKG and IKG are able to leverage their pre-trained ability to exploit the expanded modalities with knowledge passed from visibility. The resulting $\bm{\xi}_{\mathcal{M}^{v \rightarrow i},\mathcal{M}^{v \rightarrow t}|\mathcal{M}^{v}}$ can be viewed as a pool of enriched knowledge with respect to the original astronomical signals. Moreover, KB has a different instance for each sample as it is automatically updated in a sample-specific manner. Thus, we use each $\bm{\zeta}_{\mathcal{M}^{v}}$ as the query and its corresponding $\bm{\xi}_{\mathcal{M}^{v \rightarrow i},\mathcal{M}^{v \rightarrow t}|\mathcal{M}^{v}}$ as the KB and perform the feature selection and integration. Specifically, we first perform the crossmodal attention to aggregate the effective portions of KB. A single head is formulated as:
\begin{equation}
    \resizebox{.43\textwidth}{!}{$
        \bm{\kappa}^{h_j} = softmax\left( \frac{\bm{\zeta}^{h_j}_{\mathcal{M}^{v}} \cdot \bm{\xi}^{h_j}_{\mathcal{M}^{v \rightarrow i},\mathcal{M}^{v \rightarrow t}|\mathcal{M}^{v}} }{\sqrt{d_{\xi}}}^{T} \right) \bm{\xi}^{h_j}_{\mathcal{M}^{v \rightarrow i},\mathcal{M}^{v \rightarrow t}|\mathcal{M}^{v}},$}
\end{equation}
where $h_j$ is the $j$-th head of the crossmodal attention and $d_{\xi}$ is the dimension of $\bm{\xi}_{\mathcal{M}^{v \rightarrow i},\mathcal{M}^{v \rightarrow t}|\mathcal{M}^{v}}$. Then, the calculation of multimodal $\bm{\eta}_{\left\{v,i,t\right\}}$ is finalized by fusing the retrived knowledge $\bm{\kappa}$ and $\bm{\zeta}_{\mathcal{M}^{v}}$ using the residual connection as follows:
\begin{equation}
    \begin{aligned}
        \bm{\eta}_{\left\{v,i,t\right\}} = \bm{\kappa} \left( \bm{\zeta}_{\mathcal{M}^{v}} \right) + \bm{\zeta}_{\mathcal{M}^{v}}.
    \end{aligned}
\end{equation}

\begin{figure*}[t]
\centering
    \includegraphics[width=.85\linewidth]{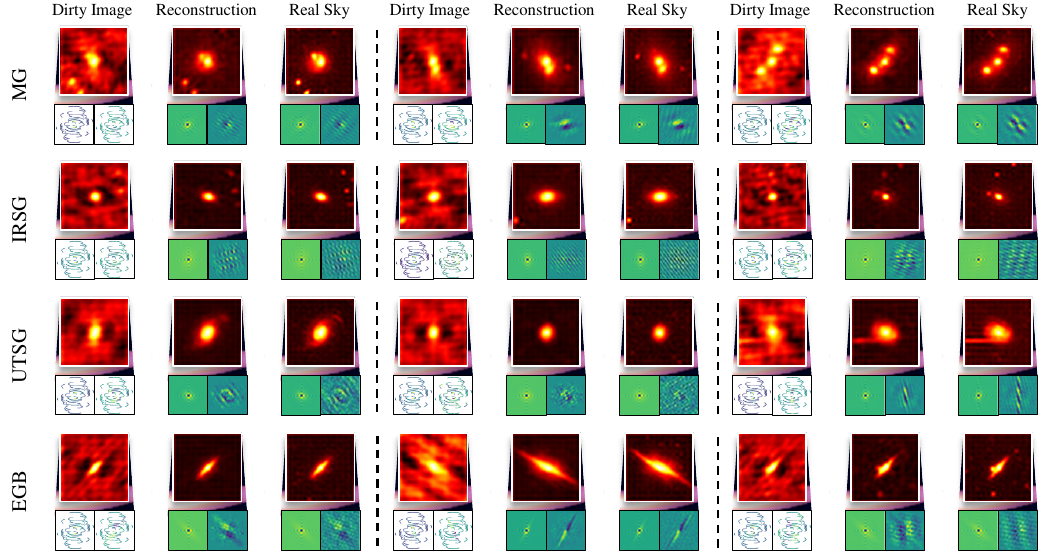}
    \caption{Visualization of the reconstruction results of VVTRec compared with the dirty image and real sky, including both visibility and image. The left half of the panel below is the real part, and the right half is the imaginary part.}
    \label{fig:fig5}
\end{figure*}

\subsection{Conditional Reconstructor}
In terms of the reconstruction process, we also only utilize the sparse visibility as the input and a neural field to fulfill the constraints of the mapping following the setting of Wu \textit{et al.} \cite{wu2022neural}. To make this process trainable together with the other modules of VVTRec, we utilize CR $\mathcal{R}_{\theta_R}$ built upon MLP to model the distribution as follows:
\begin{equation}
    \begin{aligned}
        p_{\bm{v} \sim p\left( \mathcal{M}^{v} \right), \bm{\eta} \sim p\left( \mathcal{M}^{v \rightarrow i},\mathcal{M}^{v \rightarrow t},\mathcal{M}^{v} \right)}\left( \bm{x} | \bm{v}, \bm{\eta} \right),
    \end{aligned}
\end{equation}
\begin{equation}
    \resizebox{.33\textwidth}{!}{$
    \begin{aligned}
        \hat{\mathcal{M}^{v}_d} &= \mathcal{R}_{\theta_R} \left( \mathcal{M}^{v}, \bm{\eta}_{\left\{v,i,t\right\}} \right) \\
        &=\gamma \left( \bm{\eta}^{(j)} \right) \odot \bm{\tau}_j + \beta \left( \bm{\eta}^{(j)} \right), \bm{\eta}^{(j)} \in \bm{\eta}_{\left\{v,i,t\right\}} \\
    \end{aligned}
    $},
\end{equation}
where $\bm{\tau}_j$ is the activation of $j$-th layer, $\gamma$ and $\beta$ are affine layers, and $\odot$ is the Hadamard product.

In VVTRec, the set of unfrozen trainable parameters $\Theta$ comprises $\theta_{c,1}, \theta_{c,2}, \theta_{v}, \theta_R$, ensuring that no excessive computational overhead is introduced. Thus, VVTRec is optimized by utilizing losses computed from complex values in the frequency domain \cite{jiang2021focal}:
\begin{equation}
    \begin{aligned}
        \underset{\Theta}{\min} \  \mathscr{L} \left( \hat{\mathcal{M}^{v}_d}, \mathcal{M}^{v}_d \right) = \operatorname{Avg} \left( {\omega \left|\hat{\mathcal{M}^{v}_d} - \mathcal{M}^{v}_d\right|^2} \right)
    \end{aligned},
\end{equation}

\begin{equation}
    \begin{aligned}
        \omega = \left( \frac{\rho}{\max\left( \rho \right)} + 1 \right) \left|\hat{\mathcal{M}^{v}_d} - \mathcal{M}^{v}_d\right|
    \end{aligned},
\end{equation}
where $\mathcal{M}^{v}_d$ is the ground truth, $\operatorname{Avg}$ denotes the average operation on the uv-plane, $\rho$ denotes the amplitude.

\section{Experiments}

\begin{table*}[t]
    \centering
    \caption{\textbf{Comparison results on different datasets using the full training set}. The Peak Signal to Noise Ratio (PSNR$\uparrow$: higher is better) and Structural Similarity Index Measure (SSIM$\uparrow$) are reported. \textbf{Bold} and \underline{underlined} denote the best and second best results respectively.}
\resizebox{\linewidth}{!}{
    \begin{tabular}{lc|cccccccc}
    \toprule
\multirow{2}{*}{\textbf{Method}}              & \multirow{2}{*}{\textbf{Type}}      & \multicolumn{2}{c}{\textbf{MG}} & \multicolumn{2}{c}{\textbf{IRSG}} & \multicolumn{2}{c}{\textbf{UTSG}} & \multicolumn{2}{c}{\textbf{EGB}} \\
                             &                     & PSNR$\uparrow$            & SSIM$\uparrow$            & PSNR$\uparrow$             & SSIM$\uparrow$            & PSNR$\uparrow$             & SSIM$\uparrow$            & PSNR$\uparrow$            & SSIM$\uparrow$           \\ \hline
\grow DIRTY \cite{gilbert2014recent}           & \gray{\textit{Traditional}}            & 11.633          & 0.715           & 11.398           & 0.722           & 11.890           & 0.725           & 11.356          & 0.708          \\
\grow CLEAN \cite{cornwell2008multiscale}                        & \gray{\textit{Traditional}}            & 18.557          & 0.818           & 20.913           & 0.839           & 17.048           & 0.806           & 19.448          & 0.831          \\
U-Net \cite{xie2022measurement}                      & \textit{Unimodal}            & 18.126          & 0.818           & 19.770           & 0.828           & 14.877           & 0.786           & 19.232          & 0.822          \\
Radionets \cite{schmidt2022deep}                 & \textit{Unimodal}            & 19.687          & 0.836           & 21.369           & 0.854           & 20.828           & 0.844           & 20.322          & 0.836          \\
NF \cite{wu2022neural}                  & \textit{Unimodal}           & 20.560          & 0.875           & 24.099           & 0.898           & 21.323           & 0.880           & 21.276          & 0.879        \\
VisRec \cite{wang2025visrec}                   & \textit{Unimodal}           & 22.267          & 0.883           & 24.940           & \underline{0.915}           & 23.805           & \underline{0.909}           & 23.268          & 0.901        \\
PolarRec \cite{wang2024polarrec}                        & \textit{Unimodal}           & \underline{24.088}     & \underline{0.884}           & \underline{26.234}      & 0.912           & \underline{25.384}      & 0.904           & \underline{25.420}     & \underline{0.908}          \\
\brow VVTRec(\bf{Ours})          & \blue{\textit{Multimodal}}           & \gbf{26.164}          & \dpbf{0.919}           & \gbf{28.276}           & \dpbf{0.932}           & \gbf{27.261}           & \dpbf{0.928}           & \gbf{27.061}          & \dpbf{0.924}          \\
    \bottomrule
    \end{tabular}
    }
    \label{tab:table1}
\end{table*}

\begin{table*}[t]
    \centering
    \caption{\textbf{Computational and reconstruction performance comparisons of VVTRec using different pre-trained models}. \textbf{Bold} and \underline{underlined} denote the best and second best results respectively.}
\resizebox{\linewidth}{!}{
    \begin{tabular}{l|cc|cccccccc}
    \toprule
\multirow{2}{*}{\textbf{Model}}              & \textbf{Number (M) / Memory (MiB)}    & \multirow{2}{*}{\textbf{Speed (s/iter)}}    & \multicolumn{2}{c}{\textbf{MG}} & \multicolumn{2}{c}{\textbf{IRSG}} & \multicolumn{2}{c}{\textbf{UTSG}} & \multicolumn{2}{c}{\textbf{EGB}} \\
                             & \textbf{of Learnable Parameters}          &            & PSNR$\uparrow$            & SSIM$\uparrow$            & PSNR$\uparrow$             & SSIM$\uparrow$            & PSNR$\uparrow$             & SSIM$\uparrow$            & PSNR$\uparrow$            & SSIM$\uparrow$           \\ \hline
\grow PolarRec           & 14.9 / 59.7        & 0.0094            & 24.088          & 0.884           & 26.234           & 0.912           & 25.384           & 0.904           & 25.420          & 0.908          \\
VVTRec+\textit{ViLT}              & 17.6 / 70.2        & 0.0158            & 25.747          & 0.917           & 28.134           & 0.931           & 27.192          & 0.927           & 26.764          & 0.923          \\
VVTRec+\textit{CLIP}              & 17.5 / 69.9        & 0.0138            & \gbf{26.164}          & \underline{0.919}           & \underline{28.276}           & \underline{0.932}           & \gbf{27.261}           & \underline{0.928}           & \underline{27.061}          & \underline{0.924}          \\
VVTRec+\textit{BLIP}             & 18.0 / 72.1         & 0.0623           & \underline{26.139}          & \dpbf{0.920}           & \gbf{28.363}           & \dpbf{0.933}           & \underline{27.242}           & \dpbf{0.929}           & \gbf{27.063}          & \dpbf{0.925}        \\
    \bottomrule
    \end{tabular}
    }
    \label{tab:table2}
\end{table*}

\subsection{Experimental Setup}
Following the latest methods \cite{wang2024polarrec}, we utilize a consistent telescope configuration to sample visibility data using the eht-imaging toolkit \cite{chael2018interferometric,chael2019ehtim} from real astronomical observations in different public datasets of distinct galaxy morphologies: Merging Galaxies (MG), In-between Round Smooth Galaxies (IRSG), Unbarred Tight Spiral Galaxies (UTSG), and Edge-on Galaxies with Bulge (EGB). These data are collected from the DESI Legacy Imaging Surveys \cite{dey2019overview}, integrating contributions from the Beijing-Arizona Sky Survey (BASS) \cite{zou2017project}, the DECam Legacy Survey (DE-CaLS) \cite{blum2016decam}, and the Mayall z-band Legacy Survey \cite{silva2016mayall}. The parameters for observation are adjusted to mirror an 8-telescope Event Horizon Telescope (EHT) setup to ensure consistency. We conduct experiments by utilizing the PyTorch \cite{paszke2019pytorch} framework on an NVIDIA H800 GPU. To quantitatively analyze the quality of reconstructions, we adopt two commonly used metrics, Peak Signal-to-Noise Ratio (PSNR) and Structural Similarity Index Measure (SSIM), for evaluating the generated images.

\subsection{Overall Comparison}
As shown in Table~\ref{tab:table1}, VVTRec outperforms previous methods across datasets. The best results of PSNR and SSIM are highlighted in bold, respectively. The second-best results are underlined. Overall, the average PSNR of VVTRec is 1.909 higher than the second best on four datasets. Compared with traditional methods, we can obtain an average improvement of 8.199. Such performance gains strongly show that our multimodal advancements on visibility can really contribute to the underlying feature extractions of weak signals and enhanced reconstruction. Meanwhile, improvements in SSIM show that VVTRec indeed better integrates the pre-trained knowledge with original astronomical significance to produce reconstructions with more plausible image structures than unimodal methods. This improvement enhances the reliability for astronomical research, as structural information about celestial objects often provides useful insights.

We further demonstrate the dirty image, reconstruction results of VVTRec and real sky, including both visibility and image in Figure~\ref{fig:fig5}. We can observe abundant artifacts in dirty images. VVTRec successfully reconstructs and preserves various features with different shapes of celestial objects, addressing the issue of unrealistic reconstructions caused by the point-like sources assumption. It effectively removes artifacts and noise, better supporting subsequent scientific studies.

\begin{figure}[t!]
    \centering
    \includegraphics[width=\columnwidth]{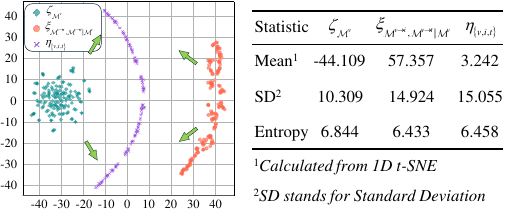}
    \caption{t-SNE visualization of the key latent components involved in the modelling of the multimodal joint representation and illustrations of their corresponding statistics.}
    \label{fig:fig6}
\end{figure}

\begin{figure*}[t]
\centering
    \includegraphics[width=.87\linewidth]{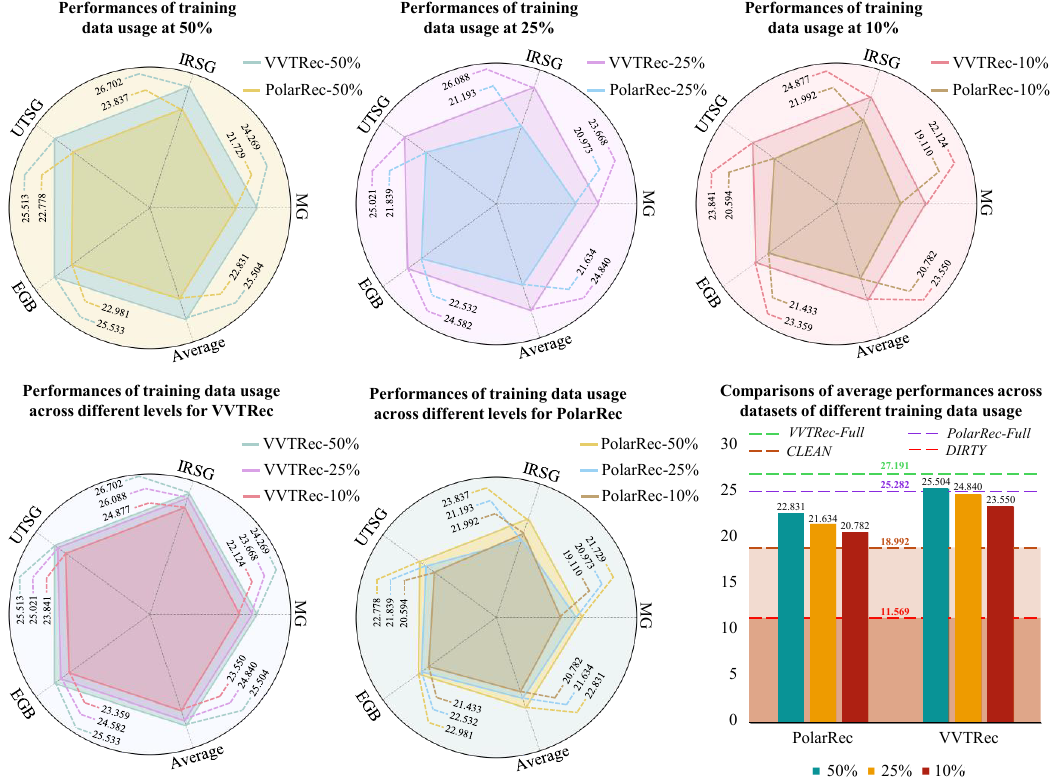}
    \caption{Radar plots comparing the PSNR results on the full test set with 10\%, 25\%, and 50\% training data usage between VVTRec and PolarRec. The bar chart further shows comparisons of results between data-efficient training and full data usage training.}
    \label{fig:fig7}
\end{figure*}

\subsection{Applicability and Computational Cost}
To demonstrate the applicability of VVTRec, we conduct extended experiments on four datasets with three representative pre-trained VLMs, namely ViLT \cite{kim2021vilt}, CLIP \cite{radford2021learning}, and BLIP \cite{li2023blip,xue2025blip} to show that boosting performances by expanding modalities in VVTRec does not necessarily introduce excessive computational overhead, as shown in Table~\ref{tab:table2}. Specifically, VVTRec+\textit{CLIP} and VVTRec+\textit{BLIP} typically gain the best and second-best results across datasets. This performance demonstrates that VVTRec can better utilize profound pre-trained knowledge. Among the three models, VVTRec+\textit{CLIP} has the fewest learnable parameters. Compared with the unimodal method PolarRec, the number of learnable parameters is only 2.6 M higher, while the memory consumption is only 10.2 MiB higher. In terms of the training speed, VVTRec+\textit{CLIP} surpasses the other two models, making it the best one balancing performance and training resources. In summary, these models all outperform previous methods, demonstrating the effectiveness and applicability of our framework. It shows that VVTRec can handle different cases of distinct VLMs providing external knowledge, and thereby aid in exploiting visibility modality deeply.

\subsection{Qualitative Analysis of Multimodal Features}
In this experiment, we visualize the key latent components $\bm{\zeta}_{\mathcal{M}^{v}}$, $\bm{\xi}_{\mathcal{M}^{v \rightarrow i},\mathcal{M}^{v \rightarrow t}|\mathcal{M}^{v}}$, and $\bm{\eta}_{\left\{v,i,t\right\}}$ to show the effective establishment of multimodal features. We utilize the t-distributed Stochastic Neighbor Embedding (t-SNE) \cite{maaten2008visualizing} to map the high-dimensional latent embeddings into a two-dimensional space as shown in Figure~\ref{fig:fig6}. We also calculate several statistics to facilitate quantitative analysis. Specifically, we conduct 1D t-SNE analysis to obtain the mean values $ \in \mathbb{R}^1$. Observing mean values, the multimodal feature $\bm{\eta}_{\left\{v,i,t\right\}}$ lies between the other two features, indicating the information fusion between the query and knowledge. Furthermore, the standard deviation of $\bm{\eta}_{\left\{v,i,t\right\}}$ is higher than that of the other two features, indicating an increased intra-class variance. Additionally, comparisons of entropy demonstrate that $\bm{\eta}_{\left\{v,i,t\right\}}$ has retained information content from both sources successfully. Qualitatively, the multimodal features in the 2D space are positioned between the other two, preserving the ring-like characteristics of $\bm{\zeta}_{\mathcal{M}^{v}}$ while also capturing the elongated features of $\bm{\xi}_{\mathcal{M}^{v \rightarrow i},\mathcal{M}^{v \rightarrow t}|\mathcal{M}^{v}}$. This phenomenon demonstrates that the information extraction from $\bm{\xi}_{\mathcal{M}^{v \rightarrow i},\mathcal{M}^{v \rightarrow t}|\mathcal{M}^{v}}$ by $\bm{\zeta}_{\mathcal{M}^{v}}$ in VVTRec is effective.

\subsection{Data-Efficient Analysis}
A data-efficient reconstruction method is particularly meaningful for astronomy, whose data acquisition through telescopes is expensive, challenging, and labor-intensive \cite{kruk2023impact}. We further conduct experiments of utilizing portions of training data but testing on the full test set to show the data efficiency of VVTRec. We choose the best unimodal method, PolarRec, as the comparison. Specifically, we first illustrate the PSNR comparisons on the full test set between VVTRec and PolarRec with 10\%, 25\%, and 50\% training data usage in the first row of Figure~\ref{fig:fig7}, respectively. Higher values (larger radius) indicate better performance. We find that VVTRec consistently outperforms PolarRec. Then, we also compare the performance of VVTRec-10\%, -25\%, and -50\% in the first radar plot of the second row. The second radar plot of the second row illustrates similar content for PolarRec. It shows that the decrease in available training data causes a smaller impact on VVTRec than PolarRec since areas shrink faster in PolarRec. At last, we show the average performance comparisons across datasets between our data-efficient training and baselines of full data usage in a bar chart. It shows that the performance of VVTRec-10\% can still surpass PolarRec-50\% even though 40\% more data has been removed. Finally, VVTRec-50\% can obtain a comparable performance with PolarRec-Full, achieving the best performance compared to unimodal methods. Overall, experimental results showcase the data efficiency of VVTRec in potential cases of scarce astronomical data.

\section{Conclusion and Future Work}
This work develops a multimodal radio interferometric data reconstruction method with visibility-guided visual and textual modality enrichment, called VVTRec. It can transform sparse visibility into expanded modalities and integrate external knowledge in a sample-specific manner to tackle the current challenges in visibility data reconstruction: (1) weak astronomical signals with insufficient reconstruction knowledge, and (2) the difficulty in supplementing knowledge for unimodal methods due to visibility data being a foreign modality. Extensive experimental results demonstrate superior performance across multiple datasets without introducing excessive computational overhead, showcasing the effectiveness of our approach. Our work provides a foundation for multimodal interferometric data reconstruction. Findings have the potential to extend beyond the immediate scope of radio astronomy. Given our method's ability to efficiently transform domain-specific features into enriched multimodal features, it holds substantial promise for deployment in other cross-disciplinary scenarios. Specifically, we envision its application in fields such as magnetic resonance imaging, as well as in seismic imaging. Future work could also involve exploring these avenues to demonstrate the versatility of our approach.

\bibliographystyle{named}
\bibliography{main}

@inproceedings{kim2021vilt,
  title={{ViLT}: Vision-and-language transformer without convolution or region supervision},
  author={Kim, Wonjae and Son, Bokyung and Kim, Ildoo},
  booktitle={International Conference on Machine Learning},
  pages={5583--5594},
  year={2021},
  organization={PMLR}
}

@inproceedings{radford2021learning,
  title={Learning transferable visual models from natural language supervision},
  author={Radford, Alec and Kim, Jong Wook and Hallacy, Chris and Ramesh, Aditya and Goh, Gabriel and Agarwal, Sandhini and Sastry, Girish and Askell, Amanda and Mishkin, Pamela and Clark, Jack and others},
  booktitle={International Conference on Machine Learning},
  pages={8748--8763},
  year={2021},
  organization={PMLR}
}

@inproceedings{li2023blip,
  title={{BLIP-2}: Bootstrapping language-image pre-training with frozen image encoders and large language models},
  author={Li, Junnan and Li, Dongxu and Savarese, Silvio and Hoi, Steven},
  booktitle={International Conference on Machine Learning},
  pages={19730--19742},
  year={2023},
  organization={PMLR}
}

@inproceedings{xue2025blip,
  title={{BLIP-3}: A family of open large multimodal models},
  author={Xue, Le and Shu, Manli and Awadalla, Anas and Wang, Jun and Yan, An and Purushwalkam, Senthil and Zhou, Honglu and Prabhu, Viraj and Dai, Yutong and Ryoo, Michael S and others},
  booktitle={Proceedings of the IEEE/CVF International Conference on Computer Vision},
  pages={6124--6135},
  year={2025}
}

@inproceedings{lin2014microsoft,
  title={Microsoft coco: Common objects in context},
  author={Lin, Tsung-Yi and Maire, Michael and Belongie, Serge and Hays, James and Perona, Pietro and Ramanan, Deva and Doll{\'a}r, Piotr and Zitnick, C Lawrence},
  booktitle={European Conference on Computer Vision},
  pages={740--755},
  year={2014},
  organization={Springer}
}

@article{gilbert2014recent,
  title={Recent developments in the sparse {Fourier} transform: A compressed {Fourier} transform for big data},
  author={Gilbert, Anna C and Indyk, Piotr and Iwen, Mark and Schmidt, Ludwig},
  journal={IEEE Signal Processing Magazine},
  volume={31},
  number={5},
  pages={91--100},
  year={2014},
  publisher={IEEE}
}

@article{cornwell2008multiscale,
  title={Multiscale {CLEAN} deconvolution of radio synthesis images},
  author={Cornwell, Tim J},
  journal={IEEE Journal of Selected Topics in Signal Processing},
  volume={2},
  number={5},
  pages={793--801},
  year={2008},
  publisher={IEEE}
}

@inproceedings{xie2022measurement,
  title={Measurement-conditioned denoising diffusion probabilistic model for under-sampled medical image reconstruction},
  author={Xie, Yutong and Li, Quanzheng},
  booktitle={International Conference on Medical Image Computing and Computer-Assisted Intervention},
  pages={655--664},
  year={2022},
  organization={Springer}
}

@article{schmidt2022deep,
  title={Deep learning-based imaging in radio interferometry},
  author={Schmidt, Kevin and Geyer, Felix and Fr{\"o}se, Stefan and Blomenkamp, P-S and Br{\"u}ggen, Marcus and De Gasperin, Francesco and Els{\"a}sser, Dominik and Rhode, Wolfgang},
  journal={Astronomy \& Astrophysics},
  volume={664},
  pages={A134},
  year={2022},
  publisher={EDP Sciences}
}

@inproceedings{wu2022neural,
  title={Neural interferometry: Image reconstruction from astronomical interferometers using transformer-conditioned neural fields},
  author={Wu, Benjamin and Liu, Chao and Eckart, Benjamin and Kautz, Jan},
  booktitle={Proceedings of the AAAI Conference on Artificial Intelligence},
  pages={2685--2693},
  year={2022}
}

@inproceedings{wang2024polarrec,
  title={{PolarRec}: Improving radio interferometric data reconstruction using polar coordinates},
  author={Wang, Ruoqi and Chen, Zhuoyang and Zhu, Jiayi and Luo, Qiong and Wang, Feng},
  booktitle={Proceedings of the IEEE/CVF Conference on Computer Vision and Pattern Recognition},
  pages={12841--12850},
  year={2024}
}

@inproceedings{ZhangGHCZ025,
  author       = {Jianke Zhang and
                  Yanjiang Guo and
                  Yucheng Hu and
                  Xiaoyu Chen and
                  Xiang Zhu and
                  Jianyu Chen},
  title        = {{UP-VLA}: A unified understanding and prediction model for embodied
                  agent},
  booktitle    = {International Conference on Machine Learning},
  year         = {2025}
}

@inproceedings{ZhaoD0GB0W25,
  author       = {Wei Zhao and
                  Pengxiang Ding and
                  Min Zhang and
                  Zhefei Gong and
                  Shuanghao Bai and
                  Han Zhao and
                  Donglin Wang},
  title        = {{VLAS}: Vision-language-action model with speech instructions for
                  customized robot manipulation},
  booktitle    = {International Conference on Learning Representations},
  year         = {2025},
}

@inproceedings{ZhongR0LW025,
  author       = {Siru Zhong and
                  Weilin Ruan and
                  Ming Jin and
                  Huan Li and
                  Qingsong Wen and
                  Yuxuan Liang},
  title        = {{Time-VLM}: Exploring Multimodal Vision-Language Models for Augmented
                  Time Series Forecasting},
  booktitle    = {International Conference on Machine Learning},
  year         = {2025},
}

@article{steyn2024h,
  title={H i galaxy signatures in the {SARAO MeerKAT Galactic Plane Survey}--{I. Probing} the richness of the great attractor wall across the inner zone of avoidance},
  author={Steyn, Nadia and Kraan-Korteweg, Ren{\'e}e C and Rajohnson, Sambatriniaina HA and Kurapati, Sushma and Chen, Hao and Frank, Bradley and Serra, Paolo and Staveley-Smith, Lister and Camilo, Fernando and Goedhart, Sharmila},
  journal={Monthly Notices of the Royal Astronomical Society: Letters},
  volume={529},
  number={1},
  pages={L88--L94},
  year={2024},
  publisher={Oxford University Press}
}

@article{xu20220,
  title={A 0.6 {Mpc H i} structure associated with {Stephan’s Quintet}},
  author={Xu, CK and Cheng, C and Appleton, PN and Duc, P-A and Gao, Y and Tang, N-Y and Yun, M and Dai, YS and Huang, J-S and Lisenfeld, U and others},
  journal={Nature},
  volume={610},
  number={7932},
  pages={461--466},
  year={2022},
  publisher={Nature Publishing Group UK London}
}

@article{chime2022sub,
  title={Sub-second periodicity in a fast radio burst},
  author={Chime/Frb Collaboration and Andersen, Bridget C and Bandura, Kevin and Bhardwaj, Mohit and Boyle, PJ and Brar, Charanjot and Breitman, Daniela and Cassanelli, Tomas and Chatterjee, Shami and Chawla, Pragya and others},
  journal={Nature},
  volume={607},
  number={7918},
  pages={256--259},
  year={2022},
  publisher={Nature Publishing Group UK London}
}

@article{bouman2018reconstructing,
  title={Reconstructing video of time-varying sources from radio interferometric measurements},
  author={Bouman, Katherine L and Johnson, Michael D and Dalca, Adrian V and Chael, Andrew A and Roelofs, Freek and Doeleman, Sheperd S and Freeman, William T},
  journal={IEEE Transactions on Computational Imaging},
  volume={4},
  number={4},
  pages={512--527},
  year={2018},
  publisher={IEEE}
}

@book{thompson2017interferometry,
  title={Interferometry and synthesis in radio astronomy},
  author={Thompson, A Richard and Moran, James M and Swenson, George W},
  year={2017},
  publisher={Springer Nature}
}

@inproceedings{wang2023conditional,
  author       = {Ruoqi Wang and
                  Zhuoyang Chen and
                  Qiong Luo and
                  Feng Wang},
  title        = {A Conditional Denoising Diffusion Probabilistic Model for Radio Interferometric Image Reconstruction},
  booktitle    = {European Conference on Artificial Intelligence},
  volume       = {372},
  pages        = {2499--2506},
  year         = {2023},
}

@inproceedings{wang2025visrec,
  title={{VisRec}: A Semi-Supervised Approach to Visibility Data Reconstruction in Radio Astronomy},
  author={Wang, Ruoqi and Wang, Haitao and Luo, Qiong and Wang, Feng and Wu, Hejun},
  booktitle={Proceedings of the AAAI Conference on Artificial Intelligence},
  volume={39},
  pages={852--860},
  year={2025}
}

@article{connor2022deep,
  title={Deep radio-interferometric imaging with {POLISH}: {DSA-2000} and weak lensing},
  author={Connor, Liam and Bouman, Katherine L and Ravi, Vikram and Hallinan, Gregg},
  journal={Monthly Notices of the Royal Astronomical Society},
  volume={514},
  number={2},
  pages={2614--2626},
  year={2022},
  publisher={Oxford University Press}
}

@inproceedings{bouman2016computational,
  title={Computational imaging for {VLBI} image reconstruction},
  author={Bouman, Katherine L and Johnson, Michael D and Zoran, Daniel and Fish, Vincent L and Doeleman, Sheperd S and Freeman, William T},
  booktitle={Proceedings of the IEEE Conference on Computer Vision and Pattern Recognition},
  pages={913--922},
  year={2016}
}

@inproceedings{jiang2021focal,
  title={Focal frequency loss for image reconstruction and synthesis},
  author={Jiang, Liming and Dai, Bo and Wu, Wayne and Loy, Chen Change},
  booktitle={Proceedings of the IEEE/CVF International Conference on Computer Vision},
  pages={13919--13929},
  year={2021}
}

@article{dey2019overview,
  title={Overview of the {DESI} legacy imaging surveys},
  author={Dey, Arjun and Schlegel, David J and Lang, Dustin and Blum, Robert and Burleigh, Kaylan and Fan, Xiaohui and Findlay, Joseph R and Finkbeiner, Doug and Herrera, David and Juneau, St{\'e}phanie and others},
  journal={The Astronomical Journal},
  volume={157},
  number={5},
  pages={168},
  year={2019},
  publisher={IOP Publishing}
}

@article{zou2017project,
  title={Project overview of the beijing--arizona sky survey},
  author={Zou, Hu and Zhou, Xu and Fan, Xiaohui and Zhang, Tianmeng and Zhou, Zhimin and Nie, Jundan and Peng, Xiyan and McGreer, Ian and Jiang, Linhua and Dey, Arjun and others},
  journal={Publications of the Astronomical Society of the Pacific},
  volume={129},
  number={976},
  pages={064101},
  year={2017},
  publisher={IOP Publishing}
}

@inproceedings{blum2016decam,
  title={The decam legacy survey},
  author={Blum, Robert D and Burleigh, Kaylan and Dey, Arjun and Schlegel, David J and Meisner, Aaron M and Levi, Michael and Myers, Adam D and Lang, Dustin and Moustakas, John and Patej, Anna and others},
  booktitle={American Astronomical Society Meeting Abstracts\# 228},
  volume={228},
  pages={317--01},
  year={2016}
}

@inproceedings{silva2016mayall,
  title={The mayall z-band legacy survey},
  author={Silva, David R and Blum, Robert D and Allen, Lori and Dey, Arjun and Schlegel, David J and Lang, Dustin and Moustakas, John and Meisner, Aaron M and Valdes, Francisco and Patej, Anna and others},
  booktitle={American Astronomical Society Meeting Abstracts\# 228},
  volume={228},
  pages={317--02},
  year={2016}
}

@article{chael2018interferometric,
  title={Interferometric imaging directly with closure phases and closure amplitudes},
  author={Chael, Andrew A and Johnson, Michael D and Bouman, Katherine L and Blackburn, Lindy L and Akiyama, Kazunori and Narayan, Ramesh},
  journal={The Astrophysical Journal},
  volume={857},
  number={1},
  pages={23},
  year={2018},
  publisher={IOP Publishing}
}

@article{chael2019ehtim,
  title={ehtim: Imaging, analysis, and simulation software for radio interferometry},
  author={Chael, Andrew A and Bouman, Katherine L and Johnson, Michael D and Narayan, Ramesh and Doeleman, Sheperd S and Wardle, John FC and Blackburn, Lindy L and Akiyama, Kazunori and Wielgus, Maciek and Chan, Chi-kwan and others},
  journal={Astrophysics Source Code Library},
  pages={ascl--1904},
  year={2019}
}

@article{paszke2019pytorch,
  title={Pytorch: An imperative style, high-performance deep learning library},
  author={Paszke, Adam and Gross, Sam and Massa, Francisco and Lerer, Adam and Bradbury, James and Chanan, Gregory and Killeen, Trevor and Lin, Zeming and Gimelshein, Natalia and Antiga, Luca and others},
  journal={Advances in Neural Information Processing Systems},
  volume={32},
  year={2019}
}

@article{maaten2008visualizing,
  title={Visualizing data using t-SNE},
  author={Maaten, Laurens van der and Hinton, Geoffrey},
  journal={Journal of Machine Learning Research},
  volume={9},
  number={Nov},
  pages={2579--2605},
  year={2008}
}

@article{kruk2023impact,
  title={The impact of satellite trails on {Hubble Space Telescope} observations},
  author={Kruk, Sandor and Garc{\'\i}a-Mart{\'\i}n, Pablo and Popescu, Marcel and Aussel, Ben and Dillmann, Steven and Perks, Megan E and Lund, Tamina and Mer{\'\i}n, Bruno and Thomson, Ross and Karadag, Samet and others},
  journal={Nature Astronomy},
  volume={7},
  number={3},
  pages={262--268},
  year={2023},
  publisher={Nature Publishing Group UK London}
}

@article{ables1974maximum,
  title={Maximum entropy spectral analysis},
  author={Ables, JG},
  journal={Astronomy and Astrophysics Supplement, Vol. 15, p. 383},
  volume={15},
  pages={383},
  year={1974}
}

@article{hogbom1974aperture,
  title={Aperture synthesis with a non-regular distribution of interferometer baselines},
  author={H{\"o}gbom, JA},
  journal={Astronomy and Astrophysics Supplement, Vol. 15, p. 417},
  volume={15},
  pages={417},
  year={1974}
}

@article{wilson1979cosmic,
  title={The cosmic microwave background radiation},
  author={Wilson, Robert W},
  journal={Science},
  volume={205},
  number={4409},
  pages={866--874},
  year={1979},
  publisher={American Association for the Advancement of Science}
}

@article{singh2022detection,
  title={On the detection of a cosmic dawn signal in the radio background},
  author={Singh, Saurabh and Nambissan T, Jishnu and Subrahmanyan, Ravi and Udaya Shankar, N and Girish, BS and Raghunathan, A and Somashekar, R and Srivani, KS and Sathyanarayana Rao, Mayuri},
  journal={Nature Astronomy},
  volume={6},
  number={5},
  pages={607--617},
  year={2022},
  publisher={Nature Publishing Group UK London}
}

@article{vaswani2017attention,
  title={Attention is all you need},
  author={Vaswani, Ashish and Shazeer, Noam and Parmar, Niki and Uszkoreit, Jakob and Jones, Llion and Gomez, Aidan N and Kaiser, {\L}ukasz and Polosukhin, Illia},
  journal={Advances in Neural Information Processing Systems},
  volume={30},
  year={2017}
}

@inproceedings{jin2025multi,
  title={A multi-view fusion approach for enhancing speech signals via short-time fractional fourier transform},
  author={Jin, Zikun and Qian, Yuhua and Liang, Xinyan and Geng, Haijun},
  booktitle={Proceedings of the International Joint Conference on Artificial Intelligence},
  pages={5508--5516},
  year={2025}
}

@inproceedings{zhao2025public,
  title={Public signaling in markets with information asymmetry using a limited number of signals},
  author={Zhao, Xu and Liu, Ren and Shen, Weiran},
  booktitle={Proceedings of the International Joint Conference on Artificial Intelligence},
  pages={4091--4099},
  year={2025}
}

\end{document}